\def\year{2016}\relax
\pdfoutput=1
\documentclass[letterpaper]{article}
\usepackage{aaai16}
\usepackage{times}
\usepackage{helvet}
\usepackage{courier}

\usepackage{graphicx}
\usepackage{amsmath}

\usepackage[ruled,vlined]{algorithm2e}
\newlength\mylen
\newcommand\myinput[1]{%
  \settowidth\mylen{\KwIn{}}
  \setlength\hangindent{\mylen}%
  \hspace*{\mylen}#1\\}

\def\argmax{\mathop{\rm arg\,max}}

\begin{document}
%
\title{Articulated Pose Estimation Using Hierarchical Exemplar-Based Models}
\author{Jiongxin Liu, Yinxiao Li, Peter Allen, Peter Belhumeur\\
Columbia University in the City of New York\\
\{liujx09, yli, allen, belhumeur\}@cs.columbia.edu
}
\maketitle
\begin{abstract}
Exemplar-based models have achieved great success on localizing the parts of semi-rigid objects. However, their efficacy on highly articulated objects such as humans is yet to be explored. Inspired by hierarchical object representation and recent application of Deep Convolutional Neural Networks (DCNNs) on human pose estimation, we propose a novel formulation that incorporates both hierarchical exemplar-based models and DCNNs in the spatial terms. Specifically, we obtain more expressive spatial models by assuming independence between exemplars at different levels in the hierarchy; we also obtain stronger spatial constraints by inferring the spatial relations between parts at the same level. As our method strikes a good balance between expressiveness and strength of spatial models, it is both effective and generalizable, achieving state-of-the-art results on different benchmarks: Leeds Sports Dataset and CUB-200-2011. 
\end{abstract}

\section{Introduction}
\noindent Articulated pose estimation from a static image remains a challenge in computer vision. The difficulty lies in the wide variations in the appearance of object due to articulated deformations. Therefore, an effective method generally relies on strong appearance models and expressive spatial models to capture the variations. More importantly, these models should be incorporated into a sensible framework where correct poses do enjoy relatively high likelihood.

There has been great progress in developing appearance and spatial models. Histogram of Gradient (HOG)~\cite{HOG05} was widely used as the part descriptor. However, HOG is rather weak and introduces visual ambiguities~\cite{HOGgles13}. Recently, Deep Convolutional Neural Networks (DCNNs) have demonstrated excellent performance in building appearance models for object detection~\cite{DNNDet13,Overfeat14,RCNNdet14} and human pose estimation~\cite{DeepPose14,CNNpart14,Alan14,JointTrain14}. Compared with a shallow classifier with hand-crafted features, DCNNs have the capacity of learning more discriminative features.

As for the spatial models, tree-structured model wins popularity by enjoying a simplified representation of object shape. In this model, parts are connected with tree edges modeled as elastic springs. On top of such model, various formulations have been proposed.~\cite{pictorial04} designed pictorial structure to combine appearance and spatial terms in a generative way. Discriminatively trained method is more powerful, and easily adapts to extended deformable part models (DPMs) with mixture of parts~\cite{Articulated_Deva11} and hierarchical representations~\cite{APM11,Interactive_part11,MODEC13,VisSym13}. There are efforts to go beyond the tree structure:~\cite{HiePoselet11} proposes a loopy model that contains composite parts implemented by Poselets~\cite{Poselet09,Poselet10};~\cite{PoseMachine14,JointTrain14} treat all the other parts as the neighbors of a target part; Grammar model has also been proposed to generalize the tree structure~\cite{Grammar13}.

As nonparametric spatial models, exemplars are also effective in part localization problems~\cite{adaptive_pose10,fiducials11,Consistency13}, as an ensemble of labeled samples (i.e., exemplars) literally capture plausible part configurations without assuming the distribution of part relations. However, the exemplar-based models have not shown much efficacy in human pose estimation due to degraded expressiveness of limited training samples~\cite{Partpair14}. 

In our work, we improve the expressiveness of exemplar-based models by leveraging hierarchy and composite parts: each composite part is treated as an object, with exemplars dictating the configuration of its child parts. By doing so, we obtain exemplar-based models covering a spectrum of granularity. In addition, we propose a novel formulation that captures the interactions between object parts in two aspects: spatial relations between parts at the same granularity level are inferred by DCNNs (inspired by~\cite{Alan14}), and constrained by exemplar-based models at that level; spatial relations between parts at different levels follow the parent-child relations in the hierarchy, which are well maintained in the bottom-up inference through exemplars. These efforts together allow us to only model the part relations in each layer via DCNNs without impairing the strength of the method. In some sense, our formulation is tailored for exemplar-based inference, which differs from other hierarchical models such as multi-layer DPMs. Also note that we use grouped parts to optimize the individual parts jointly, which differs from~\cite{Poselet09}.

\section{The Approach}
Our method features a hierarchical representation of object. We will first describe the relevant notations and introduce hierarchical exemplars. Then we will explain our formulation of pose estimation. In the end, a comparison with relevant techniques will be addressed.

\subsection{Hierarchical Exemplars}
\label{sec:hier_repr}

A hierarchical object (exemplar) contains two types of parts: \textit{atomic part} and \textit{composite part}. An atomic part $i$ is at the finest level of granularity, and can be annotated as a keypoint with pixel location $x_i$ (e.g., elbow). A composite part $k$ is the composite of its child parts (e.g., arm = \{sholder, elbow, wrist\}), and is denoted as a tight bounding box $b_k$ containing all the child parts inside. In our work, a \textit{part configuration} $X$ is denoted as the locations of atomic parts $[x_1,\ldots,x_N]$ where $N$ is the total number of atomic parts.

\begin{figure}[t]
\begin{center}
   \includegraphics[width=1\linewidth]{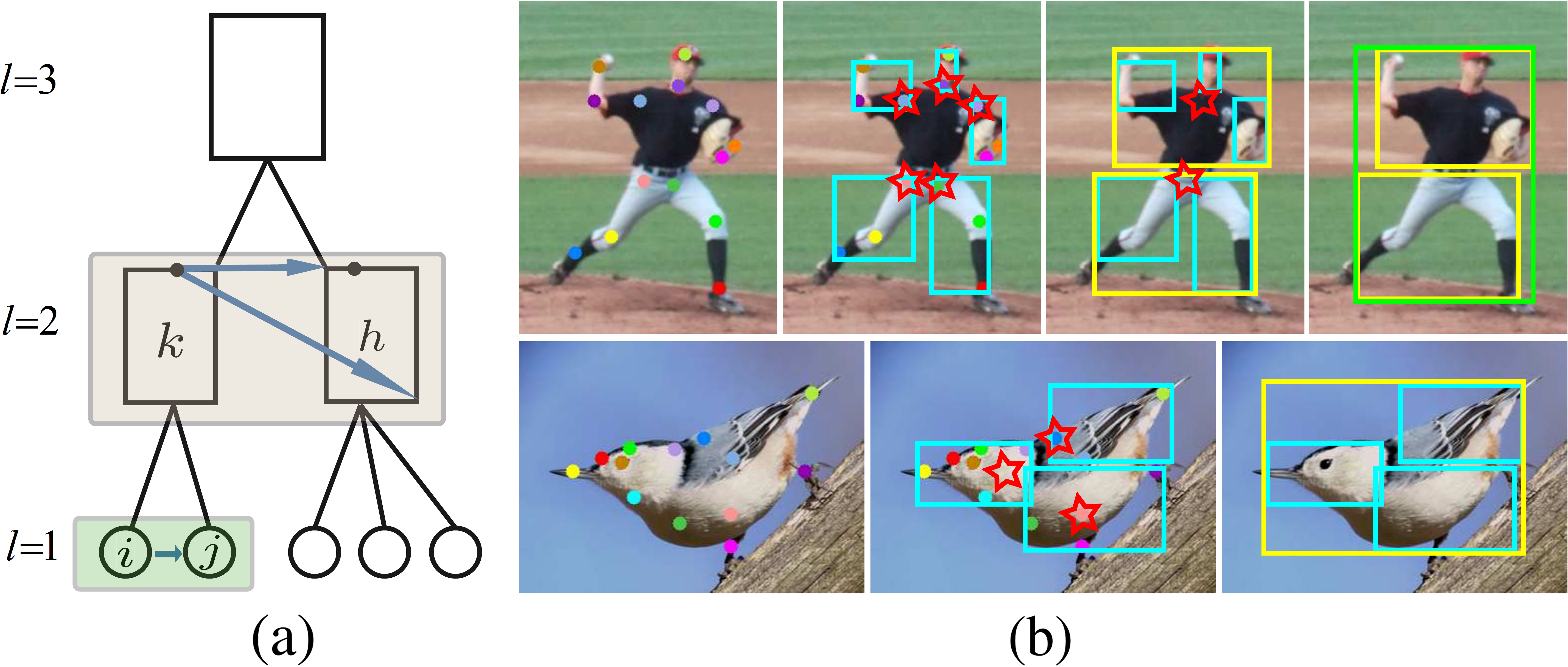}
\end{center}
   \caption{(a) shows the tree-structured hierarchy. The arrows indicate the way of estimating spatial relations (i.e., $r_{i,j}$, $r_{k,h}$) between sibling parts. The black dots on level $2$ denote the anchor points. (b) The instantiations of part hierarchy on human and bird. The part levels increase from the left to the right. Each figure shows the parts at the same level with the same color (except for the atomic parts). Immediate children are also plotted for level $2$ and above. The stars mark the anchor points for the composite parts.}
\label{fig:instant}
\end{figure}

Now, we define the \textit{spatial relation} between parts of the same type. For atomic parts $i$ and $j$, their offset $x_j-x_i$ characterizes the relation $r_{i,j}$ (e.g., shoulder is $20$ pixels above the elbow). For composite parts $k$ and $h$, we first assign anchor points $a_k$ and $a_h$ to them. Anchor points are manually determined such that they are relatively rigid w.r.t the articulated deformation. Then we represent the relation $r_{k,h}$ as $[tl(b_h)-a_k,br(b_h)-a_k]$, where $tl(\cdot)$ and $br(\cdot)$ are the top-left and bottom-right corners of part bounding box (please see Fig.~\ref{fig:instant}(a)).

The hierarchical representation follows a tree structure, as shown in Fig.~\ref{fig:instant}(a). Each leaf node denotes an atomic part. A node $k$ at level $l>1$ corresponds to a composite part $k^{(l)}$ -- the union of its children at level $l-1$ which are denoted as $C(k^{(l)})$. The degree of the tree is not bounded, and the structure of the tree depends on the particular object category. A general rule is: geometrically neighboring parts (at the same level) can form a part at an upper level if their spatial relations can be captured by the training data. Fig.~\ref{fig:instant}(b) shows the instantiation of hierarchical representation for human and bird. As the bird body is relatively more rigid than the human body, the degrees of bird's internal nodes can be larger, resulting in fewer levels.

The exemplars in previous works~\cite{fiducials11,Consistency13} correspond to a depth-2 hierarchy, where all the atomic parts are the children of the unique composite part (i.e., root part). As a result, each exemplar models the relations between all the atomic parts, making the ensemble of training exemplars not capable of capturing unseen poses. Our hierarchical exemplars, however, adapt to a hierarchy with larger depth which gives us multiple composite parts. By treating each composite part as a standalone object, we have exemplars only modeling the spatial relations between its child parts (which are referred to as part pose). We use $\mathcal{M}=\{\mathcal{M}_k^{(l)}\}|_{l>1}$ to denote the set of exemplar-based models for all the composite parts, where $k$ is the index, and $l$ denotes the level. By design, the hierarchical exemplars cover a spectrum of granularity with proper decomposition of the object, which dramatically improves the expressiveness of exemplars. Note that the depth had better not go too large, as we still want to make use of the strength of exemplars in constraining the configurations of more than two parts. 

\subsection{Formulation}
\label{sec:formulation}

We define an energy function to score a configuration $X$ given an image $I$ and the spatial models $\mathcal{M}$:
\begin{equation}
S(X|I,\mathcal{M}) = U(X|I)+R(X|I,\mathcal{M})+w_0,
\label{eq:overall_opt}
\end{equation}
\noindent where $U(\cdot)$ is the appearance term, $R(\cdot)$ is the spatial term, and $w_0$ is the bias parameter. Our goal is to find the best configuration $X^\ast=\argmax_X S(X|I,\mathcal{M})$.

\noindent \textbf{Appearance Terms:} $U(X|I)$ is a weighted combination of the detection scores for each atomic part:
\begin{equation}
U(X|I) = \sum_{i=1}^N w_i \, \varphi\left(i|I\left(x_i,s^{(1)}\left(X\right)\right)\right),
\label{eq:app_score}
\end{equation}
\noindent where $w_i$ is the weight parameter, $\varphi(\cdot)$ scores the presence of part $i$ at the location $x_i$ based on the local image patch (Eq.~\ref{eq:app_level1}), and $s^{(1)}(X)$ denotes the level-$1$ scaling factor based on $X$'s size. As we use sliding-window paradigm for detection, the local image patch is expected to fit the object size and the part's level.

\noindent\textbf{Spatial Terms:} We design multi-level spatial terms to evaluate the part relations. Assuming there are $L$ levels in the object hierarchy, and there are $n_l$ parts at the $l$-th level, then $R(X|I,\mathcal{M})$ is defined as
\begin{equation}
R(X|I,\mathcal{M}) = \sum_{l=2}^L \sum_{k=1}^{n_l} \Psi\left(p_k^{(l)}|b_k^{(l)},I,\mathcal{M}_k^{(l)}\right),
\label{eq:geo_score}
\end{equation}
\noindent where $b_k^{(l)}$ denotes the bounding box of part $k^{(l)}$, $p_k^{(l)}$ denotes the pose of $k^{(l)}$, $\mathcal{M}_k^{(l)}$ denotes the corresponding spatial models, and $\Psi(\cdot)$ scores $p_k^{(l)}$ based on both appearance and spatial models. Note that $p_k^{(l)}$ is defined as the spatial relations between the children of $k^{(l)}$.
 
We now elaborate the derivation of $\Psi(\cdot)$. Using exemplar-based models, we can assume $\mathcal{M}_k^{(l)}$ contains $T$ exemplars $\{X_i\}|_{i=1,\ldots,T}$, each of which dictates a particular pose $p_i$ (e.g., an example of raised arm). Here, we drop the subscript $k$ and superscript $l$ for clarity. With these in hand, we evaluate $\Psi(\cdot)$ as the combination of two terms:
\begin{align}
&\Psi\left(p_k^{(l)}|b_k^{(l)},I,\mathcal{M}_k^{(l)}\right) = \nonumber  \\
 &   \,\,\,         \alpha_k^{(l)} \phi\left(p_{o}|I\left(b_k^{(l)},s^{(l-1)}\left(X\right)\right)\right) + \beta_k^{(l)} \psi\left(p_k^{(l)}, p_o\right),       
\label{eq:geo_term}
\end{align}
\noindent where $\alpha_k^{(l)}$ and $\beta_k^{(l)}$ are the weight parameters, $p_o$ (corresponding to exemplar $X_o$) is the pose that best fits $p_k^{(l)}$. $\phi(\cdot)$ evaluates the likelihood of pose $p_o$ being present in the image region at $b_k^{(l)}$ (Eq.~\ref{eq:app_level2},~\ref{eq:app_level2plus}). The relevant image patches also need to be resized as Eq.~\ref{eq:app_score}. $\psi(\cdot)$ measures the similarity between $p_k^{(l)}$ and $p_o$ as
\begin{equation}
\psi(p_k^{(l)}, p_o) = -\min_t || \vec{X}_k^{(l)} - t(\vec{X}_o) ||^2,
\label{eq:psi_def}
\end{equation}
\noindent where $t$ denotes the operation of similarity transformation (the rotation angle is constrained), $\vec{X}_k^{(l)}$ denotes the vectorized locations of parts $C(k^{(l)})$ in $X$, and $\vec{X}_o$ denotes the vectorized $X_o$.

As multi-scale image cues and multi-level spatial models are both involved, $\Psi(\cdot)$ covers part relations at different levels of granularity. For instance, at a fine scale (small $l$), It evaluates whether the arm is folded; at a coarse scale (large $l$), it evaluates whether the person is seated.  

\subsection{Discussion}
\label{sec:discussion}

We make independence assumption on the spatial models in Eq.~\ref{eq:geo_score}, which can benefit articulated pose estimation. The reason lies in that it gives us a collection of spatial models that can better handle rare poses. For instance, our models allow a person to exhibit arbitrarily plausible poses at either arm as long as the spatial relations between the two arms are plausible. With such assumption, our formulation still captures the part relations thoroughly and precisely: the relations between sibling parts are encoded explicitly in Eq.~\ref{eq:geo_term}; the relations between parent and child parts are implicitly enforced (the same $X$ is referred to across the levels).

Below, we address the differences between our method and relevant techniques, such as image dependent spatial relations~\cite{Alan14,MODEC13,PosePictorial13} and hierarchical models~\cite{APM11,HiePoselet11,VisSym13}: 

\begin{itemize}
\itemsep-0.2em 
\item Unlike~\cite{Alan14,MODEC13}, our method infers from the image the spatial relations between atomic parts (e.g., elbow and shoulder), as well as the relations between composite parts (e.g.. upper body and lower body).
\item Unlike~\cite{MODEC13,PosePictorial13}, we do not conduct the selection of spatial models upfront as errors in this step are hard to correct afterwards. Instead, our selection of spatial models is based on the configuration under evaluation (the second term in Eq.~\ref{eq:geo_term}), which avoids pruning the model space too aggressively.
\item Unlike~\cite{APM11,HiePoselet11,VisSym13}, our method directly optimizes on the atomic part locations, avoiding the interference from localizing the composite parts. Also, we turn to exemplars to constrain the part relations, rather than using piece-wisely stitched ``spring models''.
\end{itemize}

\section{Inference}

The optimization of Eq.~\ref{eq:overall_opt} does not conform to general message passing framework due to the dependency of $p_o$ on $X$ (Eq.~\ref{eq:geo_term}) and the interactions between variables ${x_i}$ across multiple levels (Eq.~\ref{eq:geo_score}). Therefore, we propose an algorithm (Algorithm~\ref{alg:opt}) which simplifies the evaluation of Eq.~\ref{eq:geo_score}. Although being approximate, the algorithm is efficient and yields good results. In the following sections, we explain the two major components of the algorithm.

\subsection{Hypothesize and Test}
\label{sec:hypo_test}

The first component is \textit{Hypothesize and Test}, which leverages a RANSAC-like procedure of exemplar matching. For this, we rewrite Eq.~\ref{eq:geo_score} in a recursive form which scores the subtree rooted at $b_k^{(l)}$ ($l\ge2$):
\begin{equation}
f(b_k^{(l)}) = \sum_{j\in C(k^{(l)})}f(b_j^{(l-1)}) + \Psi\left(p_k^{(l)}|b_k^{(l)}, I,\mathcal{M}_k^{(l)}\right).
\label{eq:recur_score}
\end{equation}
\noindent Note that $f(b_k^{(1)}) = 0$ for any $k$. By comparing Eq.~\ref{eq:recur_score} with Eq.~\ref{eq:geo_score}, we can see that $f(b_1^{(L)}) = R(X|I,\mathcal{M})$.

\begin{figure}[t]
\begin{center}
   \includegraphics[width=1\linewidth]{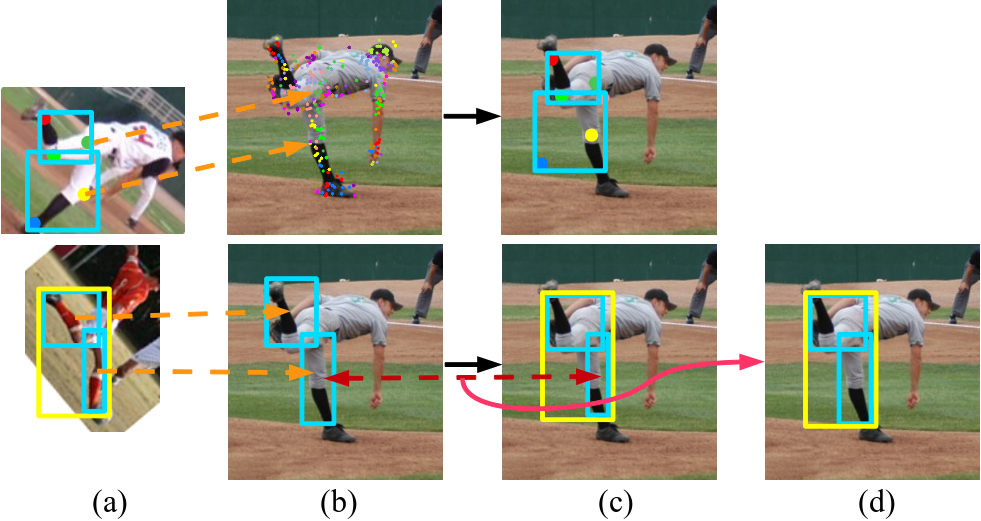}
\end{center}
   \caption{Generate hypotheses from exemplars. The first row corresponds to $l=2$ and the second row corresponds to $l=3$. (a) Two training exemplars. (b) The test image overlaid with hypotheses at level $l-1$. (c) Part hypotheses at level $l$ which are inherited from the exemplars. (d) Augmented hypothesis after swapping hypotheses of child part.}
\label{fig:algo}
\end{figure}

\textit{Hypothesize and Test} is conducted in a bottom-up manner: (1) Given the hypothesized locations of all the parts at level $l-1$ (each part has multiple hypotheses), transform the exemplars at level $l$ to the test image with similarity transformation such that each exemplar's child parts aligns with two randomly selected hypotheses of atomic parts (if $l=2$), or up to two hypotheses of composite parts (if $l>2$). (2) The geometrically well-aligned exemplars generate hypotheses for the parts at level $l$. Each hypothesis carries from exemplar the object size, the corresponding subtree, as well as the pose $p_o$ for each node in the subtree. (3) Augment the hypotheses of $k^{(l)}$ (if $l>2$) by swapping their subtrees with geometrically compatible hypotheses at level $l-1$. (4) Evaluate all the hypotheses at level $l$ using Eq.~\ref{eq:recur_score} and keep the top-scoring ones. (5) Increment $l$ and go to step (1). Fig.~\ref{fig:algo} shows examples of the first three steps.

\begin{algorithm}[t]
\small
  \caption{Inference Procedure for Pose Estimation}
  \label{alg:opt}
  \KwIn{\
		Multi-level exemplars $\{\mathcal{M}^{(l)}\}|_{l=2,\ldots,L}$\;
		\myinput{ Multi-level appearance models $\{\mathcal{C}^{(l)}\}|_{l=1,\ldots,L-1}$\;}
		\myinput{ Test image $I$;}		
		\myinput{ Maximum number of hypotheses per part Z;}}
  \KwOut{\
	The optimal configuration $X^\ast$\;}
	${hypo^{(1)}} \leftarrow$ top $Z$ local maximas from $\mathcal{C}^{(1)}(I)$, $l \leftarrow2$\;
	\While{$l \le L$} { 
	${hypo^{(l)}} \leftarrow$ randomly align $\mathcal{M}^{(l)}$ with ${hypo^{(l-1)}}$\;	
	Augment $hypo^{(l)}$ if $l>2$\;
	Evaluate $hypo^{(l)}$ using Eq.~\ref{eq:recur_score} and $\mathcal{C}^{(l-1)}(I)$\;
	${hypo^{(l)}} \leftarrow$ top-scoring $Z$ ${hypo}^{(l)}$\;
	$l \leftarrow l+1$\;
	}
	Refine and re-score $hypo^{(L)}$ through \textit{backtrack}\;
	$X^\ast \leftarrow$ highest-scoring $\hat{X}^\ast$\;
	\Return{$X^\ast$}\;
\end{algorithm}

\begin{figure}[t]
\begin{center}
\includegraphics[width=1\linewidth]{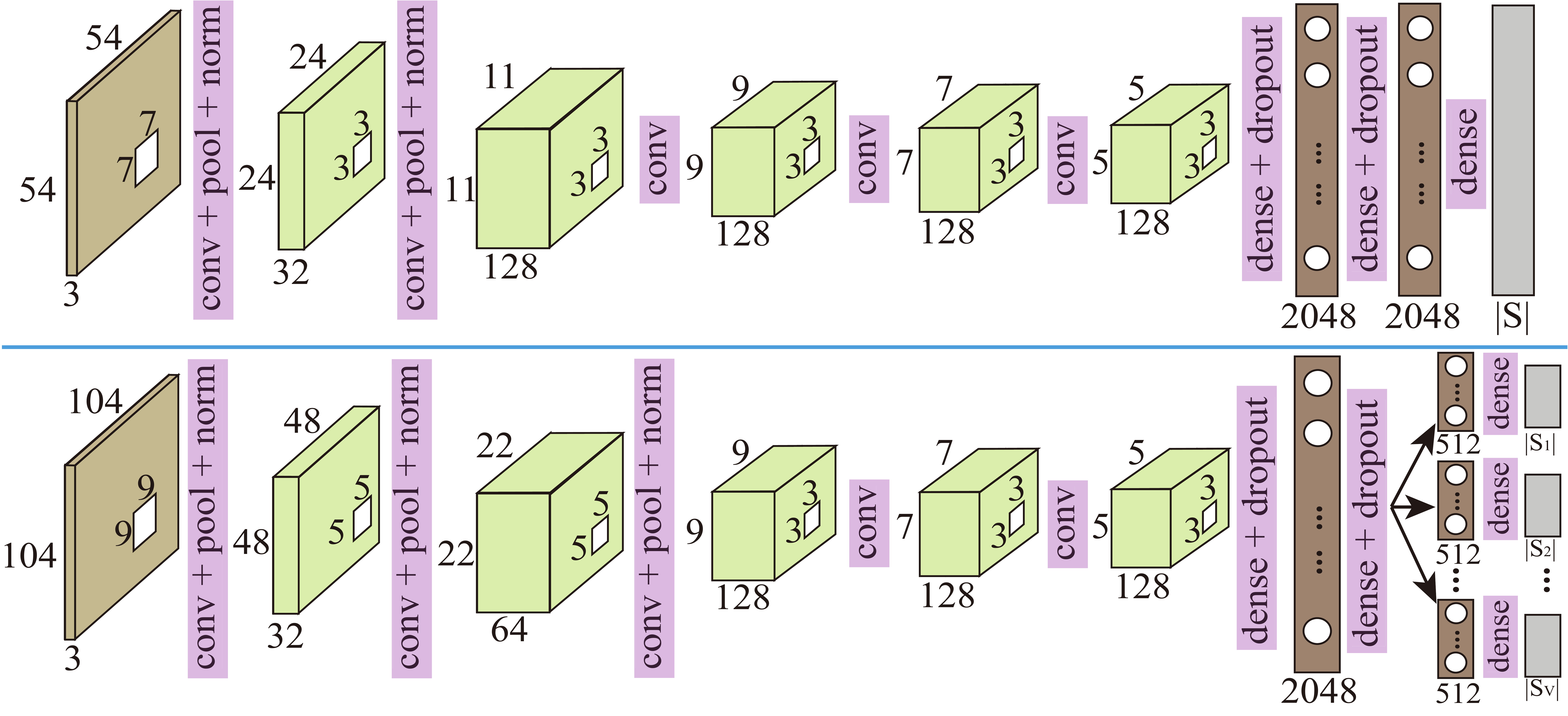}
\end{center}
\caption{{\sc{top:}} The architecture of DCNN-based atomic part detector. It consists of five convolutional layers, two max-pooling layers and three fully-connected layers. The output dimension is $|\rm{S}|$. 
{\sc{bottom:}}	The architecture of DCNN-based models for composite parts. It consists of five convolutional layers, three max-pooling layers and three fully-connected layers. The last two layers branch out, with each branch targeting the possible spatial relations of one composite part to its predefined reference part.}
\label{fig:dnn_architecture_2}
\end{figure}

\subsection{Backtrack}
The second component of the algorithm is \textit{Backtrack}. Assuming we have a hypothesis of the root part $b_1^{(L)}$, we can trace down the hierarchy, which gives us $p_o$ in Eq.~\ref{eq:geo_term} for each composite part, as well as the hypothesized locations for the atomic parts $\hat{X}=[\hat{x}_1, \hat{x}_2, \ldots, \hat{x}_N]$. 

The next step is to re-score $\hat{X}$ by obtaining its refined configuration $\hat{X}^\ast$. For this purpose, we define $g(\cdot)$ to approximate $S(\cdot)$ in Eq.~\ref{eq:overall_opt}:
\begin{equation}
g(\hat{X}|I,b_1^{(L)}) = U(\hat{X}|I) + \sum_k^{n_2} \beta_k^{(2)} \psi(p_k^{(2)}, p_o) + D, 
\label{eq:appro_S}
\end{equation}
\noindent where $D = f(b_1^{(L)}) + w_0$. Such approximation assumes $s(\hat{X})$, $p_o$ and $b_k^{(l)}$ for $l>2$ change little during the refinement. After plugging Eq.~\ref{eq:app_score} and Eq.~\ref{eq:psi_def} into Eq.~\ref{eq:appro_S}, we can solve each atomic part independently as
\begin{equation}
\hat{x}_i^\ast \! = \! \argmax_{x_i\in \mathcal{R}(\hat{x}_i)} w_i \, \varphi(i|I(x_i,s^{(1)}(\hat{X}))) + \beta_k^{(2)}||x_i-\hat{x}_i ||^2,
\label{eq:appro_opt}
\end{equation}
\noindent where part $k^{(2)}$ is the parent of part $i$, $\mathcal{R}(\hat{x}_i)$ denotes the search region of part $i$. We define the search region as a circle with radius equal to $15$\% of the average side length of $b_1^{(L)}$. We evaluate Eq. 8 for all the pixel locations inside the circle, which gives us the highest-scoring location. In the end, we obtain the refined configuration $\hat{X}^\ast=[\hat{x}_1^\ast, \hat{x}_2^\ast, \ldots, \hat{x}_N^\ast]$ with updated score $g(\hat{X}^\ast|I,b_1^{(L)})$.

\section{Model Learning}

In this section, we describe how we learn the appearance models in Eq.~\ref{eq:overall_opt} (i.e., $\varphi(\cdot)$ and $\phi(\cdot)$), as well as how we learn the weight parameters $\bf{w}$ (i.e., $w_i$, $\alpha_k^{(l)}$, and $\beta_k^{(l)}$).

\subsection{Relations Between Atomic Parts}
\label{sec:atomic_dnn}

We follow the method of~\cite{Alan14} to infer the spatial relations between atomic parts. Specifically, we design a DCNN-based multi-class classifier using Caffe~\cite{caffe14}. The architecture is shown in the first row of Fig.~\ref{fig:dnn_architecture_2}. Each value in the output corresponds to $p(i,m_{i,j}|I(x, s^{(1)}(X)))$, which is the likelihood of seeing an atomic part $i$ with a certain spatial relation (type $m_{i,j}$) to its predefined neighbor $j$, at location $x$. If $i=0$, then $m_{i,j}\in\{0\}$, indicating the background; if $i\in \{1,\ldots,N\}$, then $m_{i,j}\in\{1,\ldots,T_{i,j}\}$. By marginalization, we can derive $\varphi(\cdot)$ and $\phi(\cdot)$ as
\begin{align}
\varphi(i|I(x,s)) &= {\rm log} (p(i|I(x,s)).\label{eq:app_level1}  \\
\phi(p_o|I(b_k^{(2)},s)) &= \mkern-10mu \sum_{i\in C(k^{(2)})}\mkern-10mu{\rm log}(p(m_{i,j}|i,I(x_i,s))).
\label{eq:app_level2}
\end{align}
\noindent Note that superscript $^{(1)}$ and $X$ are dropped for clarity, $i$ and $j$ are siblings. To define type $m_{i,j}$, during training, we discretize the orientations of $r_{i,j}$ into $T_{i,j}$ (e.g., $12$) uniform bins, and $m_{i,j}$ indicates a particular bin. The training samples are then labeled as $(i,m_{i,j})$.

\subsection{Relations Between Composite Parts}
\label{sec:comp_dnn}

We build another DCNN-based model to infer the spatial relations between composite parts, as shown in the second row of Fig.~\ref{fig:dnn_architecture_2}, the architecture differs from that for atomic parts in multiple aspects. First, as the model targets composite parts which have coarser levels of granularity, the network has a larger receptive field. Second, as there are relatively fewer composite parts than atomic parts, we let all the composite parts share the features in the first several layers (the input patches of different composite parts have different receptive fields). Third, as the composite parts have different granularity with possibly significant overlap with each other, the DCNN branches out to handle them separately.

Assuming the $i$-th branch corresponds to part $i$ at level $l-1$ (Note that $l>2$), then the branch has $|{\rm S}_i|$-dim output with each value being $p(m_{i,j}|i,I(a_i, s^{(l-1)}(X)))$ based on the image patch centered at the anchor point $a_i$. Assuming the parent of part $i$ is part $k^{(l)}$, then $\phi(\cdot)$  is evaluated as
\begin{equation}
\phi(p_o|I(b_k^{(l)},s)) = \sum_{i\in C(k^{(l)})}{\rm log} (p(m_{i,j}|i,I(b_i,s))).
\label{eq:app_level2plus}
\end{equation}
\noindent Note that superscript $^{(l-1)}$ and $X$ are dropped for clarity. To train this model, we cluster the relation vector $r_{i,j}$ into $T_{i,j}$ (e.g., $24$) clusters (types) for part $i$, and the training samples are labeled accordingly.  

\subsection{Weight Parameters}
\label{sec:weight}

Eq.~\ref{eq:overall_opt} can be written as a dot product $\left\langle {\bf w}, \Phi(X,I,\mathcal{M}) \right\rangle$. Given a training sample $(X,I)$, we compute $\Phi(X,I,\mathcal{M})$ as its feature. Each training sample also has a binary label, indicating if the configuration $X$ is correct. Therefore, we build a binary max-margin classifier~\cite{S_SVM04} to estimate ${\bf w}$, with non-negativity constraints imposed. To avoid over-fitting, the training is conducted on a held-out validation set that was not used to train the DCNNs.

Before training, we augment the positive samples by randomly perturbing their part locations as long as they are reasonably close to the ground-truth locations. To generate the negative samples, we randomly place the configurations of positive samples at the incorrect regions of the training images, with Gaussian noise added to the part locations.

\begin{figure}[!htpb]
\begin{center}
\includegraphics[width=1\linewidth]{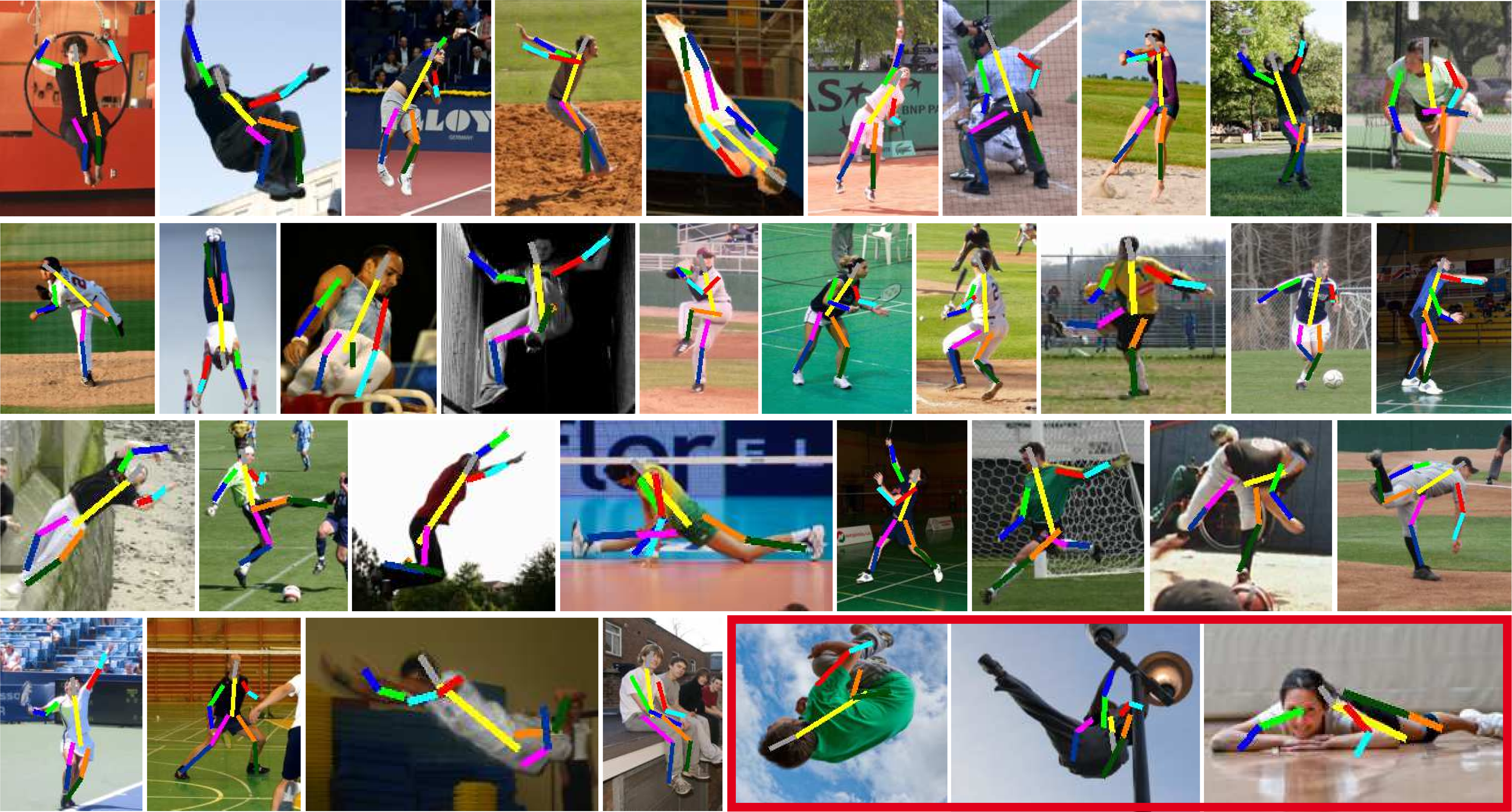}
\end{center}
   \caption{Qualitative results of human pose estimation on LSP dataset (OC annotations). Failures are denoted with red frames, which are due to extreme self-occlusion.}
\label{fig:final_results}
\end{figure}

\begin{figure}[t]
\begin{center}
\includegraphics[width=1\linewidth]{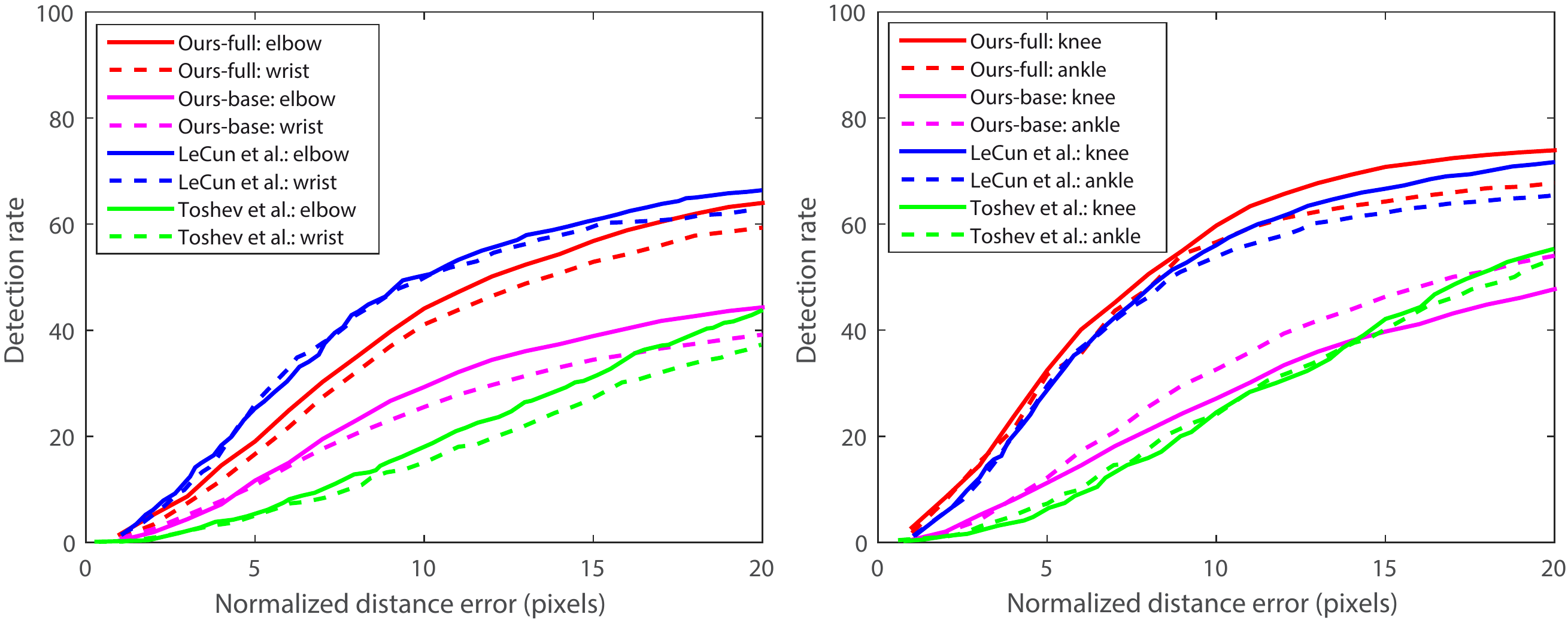}
\end{center}
   \caption{Detection rate vs. normalized error curves on the LSP Extended dataset (PC Annotations).
	{\sc{left:}} arm (elbow and wrist) detection. {\sc{right:}} leg (knee and ankle) detection.}
\label{fig:model_performance}
\end{figure}

\section{Experiments}

We evaluate our method extensively on multiple benchmarks, and conduct diagnostic experiments to show the effect of different components in our method. 

\subsection{LSP Dataset (OC Annotations)}
\label{sec:lsp}

The Leeds Sports Pose (LSP) dataset~\cite{LSP10} includes $1,000$ images for training and $1,000$ images for testing, where each image is annotated with $14$ joint locations. We augment the training data by left-right flipping, and rotating through $360^\circ$. We use observer-centric (OC) annotations to have fair comparisons with the majority of existing methods. To measure the performance, we use Percentage of Correct Parts (PCP). In PCP measure, a ``part'' is defined as a line segment connecting two neighboring joints. If both of the segment endpoints (joints) lie within $50$\% of the length of the ground-truth annotated endpoints, then the part is correct.

In this experiment, we build a hierarchy of four levels for human body. The first level contains the atomic body joints; the second level has five composite parts (Head, Right arm, Left arm, Right leg, and Left leg); the third level has two composite parts (Head\&Arms and Legs); and the fourth level corresponds to the whole body. To gain an understanding of the effect of the components of our inference algorithm, we evaluate our full method (which will be referred to as ``Ours-full''), and variants of our method (which will be referred to as ``Ours-partial'', and ``Ours-no-HIER''). Ours-full corresponds to the whole inference algorithm; Ours-partial only conducts the first part of the inference algorithm, traces down the best root hypothesis based on Eq.~\ref{eq:recur_score}, and outputs the locations of its atomic parts; Ours-no-HIER only uses full-body exemplars (after augmentation) as the spatial models.

The quantitative results of our method as well as its counterparts are listed in Tab.~\ref{tbl:LSP}. Ours-full generally outperforms the state-of-the-art methods on all the parts. The improvement over IDPR~\cite{Alan14} demonstrates the effect of reasoning multi-level spatial relations. We expect to see even larger improvement if we augment the annotations with midway points between joints. We also experiment with person-centric (PC) annotations on the same dataset, where the accuracy drops slightly. Ours-full achieves improvement over Ours-partial and Ours-no-HIER by a large margin, which demonstrates the benefits of \textit{backtrack} (higher precision) and hierarchical exemplars (more expressive models). Note that Ours-partial already outperforms Strong-PS~\cite{Strong13} and PoseMachine~\cite{PoseMachine14}.

Fig.~\ref{fig:final_results} shows some testing examples, which are selected with high diversity in poses. We can see that our method achieves accurate localization for most of the body joints, even in the case of large articulated deformation.

\begin{table}[t]
\footnotesize
\begin{center}
\setlength{\tabcolsep}{.3em}
\begin{tabular}{l|c|c|c|c|c|c|c}
Method & Torso & ULeg & LLeg & UArm & LArm & Head & Avg \\
\hline
Strong-PS & 88.7 & 78.8 & 73.4 & 61.5 & 44.9 & 85.6 & 69.2 \\
PoseMachine & 88.1 & 78.9 & 73.4 & 62.3 & 39.1 & 80.9 & 67.6 \\
IDPR & 92.7 & 82.9 & 77.0 & 69.2 & {\bf 55.4} & 87.8 & 75.0 \\
Ours-partial & 89.2 & 79.5 & 73.6 & 65.8 & 50.3 & 85.6 & 71.3 \\
Ours-no-HIER & 85.4 & 75.3 & 66.7 & 54.9 & 37.5 & 82.5 & 63.7 \\
Ours-full & 93.5 & {\bf 84.4} & {\bf 78.3} & {\bf 71.4} & 55.2 & {\bf 88.6} & {\bf 76.1} \\
Ours-full (PC) & {\bf 93.7} & 82.2 & 76.0 & 68.6 & 53.2 & 88.3 & 74.2 \\
\hline
\end{tabular}
\end{center}
\caption{Comparison of pose estimation results (PCP) on LSP dataset. Our method achieves the best performance.}
\label{tbl:LSP}
\end{table}

\subsection{LSP Extended Dataset (PC Annotations)}

To have fair comparisons with~\cite{DeepPose14,JointTrain14}, we train and test our models on LSP extended dataset using PC annotations. Altogether, we have $11,000$ training images and $1000$ testing images. As the quality of the annotations for the additional training images varies a lot, we manually filter out about $20\%$ of them. We also augment the training data through flipping and rotation.

\begin{table*}[t]
\footnotesize
\begin{center}
\begin{tabular}{l|c|c|c|c|c|c|c|c|c|c|c|c|c|c|c|c}
    Method & Ba & Bk & Be & Br & Cr & Fh & Le & Ll & Lw & Na & Re & Rl & Rw & Ta & Th & Total \\
\hline
CoE & 62.1 & 49.0 & 69.0 & 67.0 & 72.9 & 58.5 & 55.8 & 40.9 & 71.6 & 70.8 & 55.5 & 40.5 & 71.6 & 40.2 & 70.8 & 59.7 \\
Part-pair & 64.5 & 61.2 & 71.7 & 70.5 & 76.8 & 72.0 & 69.8 & 45.0 & 74.3 & 79.3 & 70.1 & 44.9 & 74.4 & 46.2 & 80.0 & 66.7 \\
DCNN-CoE & 64.7 & 63.1 & 74.2 & 71.6 & 76.3 & 72.9 & 69.0 & 48.2 & 72.6 & 82.0 & 69.2 & 47.9 & 72.3 & 46.8 & 81.5 & 67.5 \\
Ours-partial & 65.1 & 64.2 & 74.6 & 72.4 & 77.1 & 73.8 & 70.2 & 48.4 & 73.2 & 82.5 & 70.6 & 48.7 & 73.0 & 48.3 & 82.2 & 68.3 \\
Ours-full & {\bf 67.3} & {\bf 65.6} & {\bf 75.9} & {\bf 74.4} & {\bf 78.8} & {\bf 75.3} & {\bf 72.5} & {\bf 50.9} & {\bf 75.4} & {\bf 84.7}  & {\bf 72.8} & {\bf 50.4} & {\bf 75.2} & {\bf 49.9} & {\bf 84.2} & {\bf 70.2} \\
\hline
\end{tabular} 
\end{center}
\caption{Comparison of part localization results on the CUB-200-2011 bird dataset. Our method outperforms the previous methods by a large margin. From left to right, the parts are: Back, Beak, Belly, Breast, Crown, Forehead, Left Eye, Left Leg, Left Wing, Nape, Right Eye, Right Leg, Right Wing, Tail, Throat, and Total.}
\label{table:global_loc}
\end{table*}

We use Percentage of Detected Joints (PDJ) to evaluate the performance, which provides an informative view of the localization precision. In this experiment, we evaluate the baseline of our method (referred to as ``Ours-base'') by only using the first term in Eq.~\ref{eq:overall_opt}. It is equivalent to localizing the parts independently. In Fig.~\ref{fig:model_performance}, we plot the detection rate vs. normalized error curves for different methods. From the curves, we can see that Ours-base already achieves better accuracy than~\cite{DeepPose14} except for Knee. It demonstrates that a detector that scores the part appearance is more effective than a regressor that predicts the part offset. Ours-full achieves significant improvement over Ours-base by incorporating the multi-level spatial models. Our method is also comparable to~\cite{JointTrain14} which enjoys the benefit of jointly learning appearance models and spatial context.~\cite{JointTrain14} has higher accuracy on the lower arms, while we have better results on the lower legs. Also note that~\cite{JointTrain14} requires delicate implementation of a sophisticated network architecture, while our method allows the use of off-the-shelf DCNN models.

\begin{figure}[!htpb]
\begin{center}
\includegraphics[width=1\linewidth]{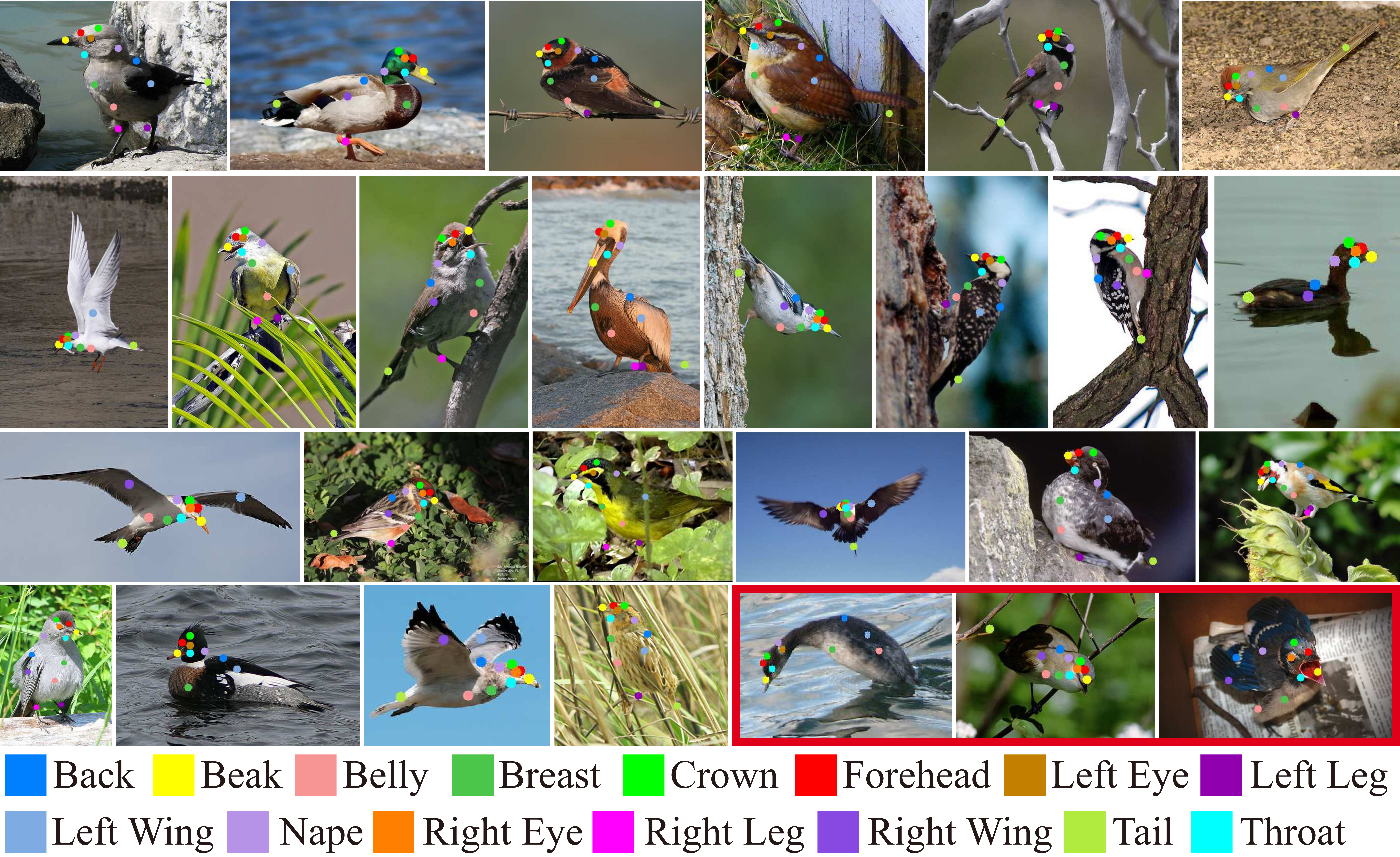}
\end{center}
   \caption{Qualitative results of part localization on CUB-200-2011 bird dataset. Failures are denoted with red frames, where some parts dramatically deviate from the correct locations due to large deformation and noisy patterns. The color codes are shown at the bottom.}
\label{fig:final_results_bird}
\end{figure}

\subsection{CUB-200-2011 Bird Dataset}

We also evaluate our method on the CUB-200-2011 bird dataset, which contains $5,994$ images for training and $5,794$ images for testing. Each image is annotated with image locations for $15$ parts. We also augment the training data through flipping and rotation. As birds are less articulated than humans, we design a three-level hierarchy for birds. The first level contains the atomic parts; the second level has three composite parts (Head, Belly\&Legs, and Back\&Tail); and the third level corresponds to the whole bird. Although we did not prove that the manually-designed hierarchy is optimal, we empirically find that it facilitates the prediction of part relations.

We use PCP to measure performance. In the bird dataset, a correct part detection should be within $1.5$ standard deviation of an MTurk worker's click from the ground-truth location. For a semi-rigid object such as bird with sufficient training samples, directly applying exemplar-based models can produce very good results. Therefore, we replace the part detectors in~\cite{Consistency13} with DCNN-based detectors (only targeting the atomic parts), which will be referred to as ``DCNN-CoE''.

We compare the results of different methods in Tab.~\ref{table:global_loc}, including CoE~\cite{Consistency13} and Part-pair~\cite{Partpair14}. First, DCNN-CoE outperforms CoE significantly, demonstrating that DCNN is much more powerful than the conventional classification model (e.g., SVM). DCNN-CoE also outperforms Part-pair with much less overhead, thanks to the efficiency of multi-class detector. Using our new method, the localization accuracy is further improved. Ours-partial improves slightly over DCNN-CoE, which is reasonable as Ours-partial is essentially multi-level DCNNs plus multi-level exemplars, and the flexibility from our multi-level exemplars has limited effect for semi-rigid objects. Also note that Ours-partial uses an incomplete scoring function. By considering the full scoring function, Ours-full achieves the best results on all the parts.

Some qualitative results are shown in Fig.~\ref{fig:final_results_bird}. From the examples, we can see that our method is capable of capturing a wide range of poses, shapes and viewpoints. In addition, our method localizes the bird parts with very high precision.

\section{Conclusion}

In this paper, we propose a novel approach for articulated pose estimation. The approach exploits the part relations at different levels of granularity through multi-scale DCNN-based models and hierarchical exemplar-based models. By incorporating DCNN-based appearance models in the spatial terms, our method couples spatial models with them, thus better adapting to the particular test image than otherwise. By introducing hierarchy in the exemplar-based models, we enjoy much more expressive spatial models even if the training data are limited. In addition, We propose an efficient algorithm to infer ``good-enough'' part configurations from a less simplified formulation. These efforts together enable us to achieve state-of-the-art results on different datasets, which demonstrates the effectiveness and generalization ability of our method.

\section{ Acknowledgments}
This work was supported by NSF awards 0968546 and 1116631, ONR award N00014-08-1-0638, and Gordon and Betty Moore Foundation grant 2987.

\bibliographystyle{aaai}

\begin{thebibliography}{}

\bibitem[\protect\citeauthoryear{Belhumeur \bgroup et al\mbox.\egroup
  }{2011}]{fiducials11}
Belhumeur, P.~N.; Jacobs, D.~W.; Kriegman, D.~J.; and Kumar, N.
\newblock 2011.
\newblock Localizing parts of faces using a consensus of exemplars.
\newblock {\em Proc. CVPR}.

\bibitem[\protect\citeauthoryear{Bourdev and Malik}{2009}]{Poselet09}
Bourdev, L., and Malik, J.
\newblock 2009.
\newblock Poselets: Body part detectors trained using 3d human pose
  annotations.
\newblock {\em Proc. ICCV}.

\bibitem[\protect\citeauthoryear{Bourdev \bgroup et al\mbox.\egroup
  }{2010}]{Poselet10}
Bourdev, L.; Maji, S.; Brox, T.; and Malik, J.
\newblock 2010.
\newblock Detecting people using mutually consistent poselet activations.
\newblock {\em Proc. ECCV}.

\bibitem[\protect\citeauthoryear{Branson, Perona, and
  Belongie}{2011}]{Interactive_part11}
Branson, S.; Perona, P.; and Belongie, S.
\newblock 2011.
\newblock Strong supervision from weak annotation: Interactive training of
  deformable part models.
\newblock {\em Proc. ICCV}.

\bibitem[\protect\citeauthoryear{Chen and Yuille}{2014}]{Alan14}
Chen, X., and Yuille, A.
\newblock 2014.
\newblock Articulated pose estimation by a graphical model with image dependent
  pairwise relations.
\newblock {\em Proc. NIPS}.

\bibitem[\protect\citeauthoryear{Dalal and Triggs}{2005}]{HOG05}
Dalal, N., and Triggs, B.
\newblock 2005.
\newblock Histograms of oriented gradients for human detection.
\newblock {\em Proc. CVPR} 1:886--893.

\bibitem[\protect\citeauthoryear{Felzenszwalb and
  Huttenlocher}{2005}]{pictorial04}
Felzenszwalb, P.~F., and Huttenlocher, D.~P.
\newblock 2005.
\newblock Pictorial structures for object recognition.
\newblock {\em IJCV} 61(1):55--79.

\bibitem[\protect\citeauthoryear{Girshick \bgroup et al\mbox.\egroup
  }{2014}]{RCNNdet14}
Girshick, R.; Donahue, J.; Darrell, T.; and Malik, J.
\newblock 2014.
\newblock Rich feature hierarchies for accurate object detection and semantic
  segmentation.
\newblock {\em Proc. CVPR}.

\bibitem[\protect\citeauthoryear{Jain \bgroup et al\mbox.\egroup
  }{2014}]{CNNpart14}
Jain, A.; Tompson, J.; Andriluka, M.; Taylor, G.~W.; and Bregler, C.
\newblock 2014.
\newblock Learning human pose estimation features with convolutional networks.
\newblock {\em Proc. ICLR}.

\bibitem[\protect\citeauthoryear{Jia \bgroup et al\mbox.\egroup
  }{2014}]{caffe14}
Jia, Y.; Shelhamer, E.; Donahue, J.; Karayev, S.; Long, J.; Girshick, R.;
  Guadarrama, S.; and Darrell, T.
\newblock 2014.
\newblock {C}affe: {C}onvolutional architecture for fast feature embedding.
\newblock {\em Proc. MM}.

\bibitem[\protect\citeauthoryear{Johnson and Everingham}{2010}]{LSP10}
Johnson, S., and Everingham, M.
\newblock 2010.
\newblock Clustered pose and nonlinear appearance models for human pose
  estimation.
\newblock {\em Proc. BMVC}.

\bibitem[\protect\citeauthoryear{Liu and Belhumeur}{2013}]{Consistency13}
Liu, J., and Belhumeur, P.~N.
\newblock 2013.
\newblock Bird part localization using exemplar-based models with enforced pose
  and subcategory consistency.
\newblock {\em Proc. ICCV}.

\bibitem[\protect\citeauthoryear{Liu, Li, and Belhumeur}{2014}]{Partpair14}
Liu, J.; Li, Y.; and Belhumeur, P.~N.
\newblock 2014.
\newblock Part-pair representation for part localization.
\newblock {\em Proc. ECCV}.

\bibitem[\protect\citeauthoryear{Pishchulin \bgroup et al\mbox.\egroup
  }{2013a}]{PosePictorial13}
Pishchulin, L.; Andriluka, M.; Gehler, P.; and Schiele, B.
\newblock 2013a.
\newblock Poselet conditioned pictorial structures.
\newblock {\em Proc. CVPR}.

\bibitem[\protect\citeauthoryear{Pishchulin \bgroup et al\mbox.\egroup
  }{2013b}]{Strong13}
Pishchulin, L.; Andriluka, M.; Gehler, P.; and Schiele, B.
\newblock 2013b.
\newblock Strong appearance and expressive spatial models for human pose
  estimation.
\newblock {\em Proc. ICCV}.

\bibitem[\protect\citeauthoryear{Ramakrishna \bgroup et al\mbox.\egroup
  }{2014}]{PoseMachine14}
Ramakrishna, V.; Munoz, D.; Hebert, M.; Bagnell, J.~A.; and Sheikh, Y.
\newblock 2014.
\newblock Pose machines: Articulated pose estimation via inference machines.
\newblock {\em Proc. ECCV}.

\bibitem[\protect\citeauthoryear{Rothrock, Park, and Zhu}{2013}]{Grammar13}
Rothrock, B.; Park, S.; and Zhu, S.-C.
\newblock 2013.
\newblock Integrating grammar and segmentation for human pose estimation.
\newblock {\em Proc. CVPR}.

\bibitem[\protect\citeauthoryear{Sapp and Taskar}{2013}]{MODEC13}
Sapp, B., and Taskar, B.
\newblock 2013.
\newblock Modec: Multimodal decomposable models for human pose estimation.
\newblock {\em Proc. CVPR}.

\bibitem[\protect\citeauthoryear{Sapp, Jordan, and
  Taskar}{2010}]{adaptive_pose10}
Sapp, B.; Jordan, C.; and Taskar, B.
\newblock 2010.
\newblock Adaptive pose priors for pictorial structures.
\newblock {\em Proc. CVPR}.

\bibitem[\protect\citeauthoryear{Sermanet \bgroup et al\mbox.\egroup
  }{2014}]{Overfeat14}
Sermanet, P.; Eigen, D.; Zhang, X.; Mathieu, M.; Fergus, R.; and LeCun, Y.
\newblock 2014.
\newblock Overfeat: Integrated recognition, localization and detection using
  convolutional networks.
\newblock {\em Proc. ICLR}.

\bibitem[\protect\citeauthoryear{Sun and Savarese}{2011}]{APM11}
Sun, M., and Savarese, S.
\newblock 2011.
\newblock Articulated part-based model for joint object detection and pose
  estimation.
\newblock {\em Proc. ICCV}.

\bibitem[\protect\citeauthoryear{Szegedy, Toshev, and Erhan}{2013}]{DNNDet13}
Szegedy, C.; Toshev, A.; and Erhan, D.
\newblock 2013.
\newblock Deep neural networks for object detection.
\newblock {\em Proc. NIPS}.

\bibitem[\protect\citeauthoryear{Tompson \bgroup et al\mbox.\egroup
  }{2014}]{JointTrain14}
Tompson, J.; Jain, A.; LeCun, Y.; and Bregler, C.
\newblock 2014.
\newblock Joint training of a convolutional network and a graphical model for
  human pose estimation.
\newblock {\em Proc. NIPS}.

\bibitem[\protect\citeauthoryear{Toshev and Szegedy}{2014}]{DeepPose14}
Toshev, A., and Szegedy, C.
\newblock 2014.
\newblock Deeppose: Human pose estimation via deep neural networks.
\newblock {\em Proc. CVPR}.

\bibitem[\protect\citeauthoryear{Tsochantaridis \bgroup et al\mbox.\egroup
  }{2004}]{S_SVM04}
Tsochantaridis, I.; Hofmann, T.; Joachims, T.; and Altun, Y.
\newblock 2004.
\newblock Support vector machine learning for interdependent and structured
  output spaces.
\newblock {\em Proc. ICML}.

\bibitem[\protect\citeauthoryear{Vondrick \bgroup et al\mbox.\egroup
  }{2013}]{HOGgles13}
Vondrick, C.; Khosla, A.; Malisiewicz, T.; and Torralba, A.
\newblock 2013.
\newblock Hoggles: Visualizing object detection features.
\newblock {\em Proc. ICCV}.

\bibitem[\protect\citeauthoryear{Wang and Li}{2013}]{VisSym13}
Wang, F., and Li, Y.
\newblock 2013.
\newblock Learning visual symbols for parsing human poses in images.
\newblock {\em Proc. IJCAI}.

\bibitem[\protect\citeauthoryear{Wang, Tran, and Liao}{2011}]{HiePoselet11}
Wang, Y.; Tran, D.; and Liao, Z.
\newblock 2011.
\newblock Learning hierarchical poselets for human parsing.
\newblock {\em Proc. CVPR}.

\bibitem[\protect\citeauthoryear{Yang and Ramanan}{2011}]{Articulated_Deva11}
Yang, Y., and Ramanan, D.
\newblock 2011.
\newblock Articulated pose estimation with flexible mixtures-of-parts.
\newblock {\em Proc. CVPR}.

\end{thebibliography}

\end{document}